\documentclass[sigconf]{acmart}

\makeatletter
\def\@ACM@checkaffil{%
  \if@ACM@instpresent\else\ClassWarningNoLine{\@classname}{No institution present}\fi
  \if@ACM@citypresent\else\ClassWarningNoLine{\@classname}{No city present}\fi
  \if@ACM@countrypresent\else\ClassWarningNoLine{\@classname}{No country present}\fi}
\makeatother

\AtBeginDocument{%
  }

\acmYear{2025}
\acmConference[SC'25]{Supercomputing ’25}{November 16--21, 2025}{St. Louis, MO, USA}
\acmBooktitle{Supercomputing’25 (SC '25), November 16--21, 2025, St. Louis, MO, USA}

\usepackage{amsmath,amsfonts}
\usepackage{algorithmic}
\usepackage{graphicx}
\usepackage{xcolor}
\usepackage{tabularx}
\usepackage[detect-all,group-separator={,}]{siunitx}
\usepackage{url}
\usepackage{enumitem}
\usepackage{textcomp}
\usepackage{bibentry}
\usepackage{natbib}
\usepackage{color}
\usepackage{xspace}
\usepackage{bigstrut}
\usepackage{multirow}
\usepackage{subcaption}
\usepackage{balance}
\usepackage{makecell}
\usepackage{hyperref}
\usepackage{float}
\usepackage{caption}

\begin{document}


\title[Automated MCQA Benchmarking at Scale]{Automated MCQA Benchmarking at Scale: Evaluating Reasoning Traces as Retrieval Sources for Domain Adaptation of Small Language Models}


\author{
Ozan Gokdemir$^{1,2^{*}}$,
Neil Getty$^{1}$,
Robert Underwood$^{1}$,
Sandeep Madireddy$^{1}$,
Franck Cappello$^{1}$,
Arvind Ramanathan$^{1,2}$,
Ian T. Foster$^{1,2}$,
Rick L. Stevens$^{1,2}$
}

\affiliation{
 $^{1}$Argonne National Laboratory, Lemont, Illinois, USA 
 $^{2}$The University of Chicago, Chicago, Illinois, USA 
}

\affiliation{ $^{*}$Corresponding author, ogokdemir@anl.gov, ogokdemir@uchicago.edu}

\renewcommand{\shortauthors}{Gokdemir, et al.}

\begin{abstract}\
As scientific knowledge grows at an unprecedented pace, evaluation benchmarks must evolve to reflect new discoveries and ensure language models are tested on current, diverse literature. We propose a scalable, modular framework for generating multiple-choice question-answering (MCQA) benchmarks directly from large corpora of scientific papers. Our pipeline automates every stage of MCQA creation, including PDF parsing, semantic chunking, question generation, and model evaluation. As a case study, we generate more than 16,000 MCQs from 22,000 open-access articles in radiation and cancer biology. We then evaluate a suite of small language models (1.1B–14B parameters) on these questions, comparing baseline accuracy with retrieval-augmented generation (RAG) from paper-derived semantic chunks and from reasoning traces distilled from GPT-4.1. We find that reasoning-trace retrieval consistently improves performance on both synthetic and expert-annotated benchmarks, enabling several small models to surpass GPT-4 on the 2023 Astro Radiation and Cancer Biology exam.

\end{abstract}


\maketitle

\section{Introduction}
In the rapidly evolving landscape of machine learning, benchmarks have been central to progress, providing standardized tests to compare models and track improvements \cite{rein2023gpqagraduatelevelgoogleproofqa,hendrycks2021mmlu}. However, as scientific knowledge expands at an unprecedented pace \cite{landhuis2016, nsb2024}, existing benchmarks struggle to keep up with new discoveries and ensure that language models (LLMs) are tested on current, diverse, and domain-relevant literature. Widely used multiple-choice question answering (MCQA) benchmarks such as ARC \cite{clark2018think}, GPQA \cite{rein2023gpqagraduatelevelgoogleproofqa}, SciQA \cite{wang2024llmsforSciQA}, PaperQA \cite{Lala2023PaperQA}, and MMLU \cite{hendrycks2021mmlu} have driven advances in reasoning, but they are largely static, often limited to general knowledge, and increasingly prone to contamination by pretraining corpora. At the same time, manually curating domain-specific benchmarks is prohibitively expensive, slow to update, and difficult to reproduce at scale.

To address these limitations, we introduce a scalable and modular framework for automated MCQA benchmark generation from large scientific corpora. Our framework automates the end-to-end workflow, including PDF parsing, semantic chunking, question and distractor generation, and quality control, while maintaining provenance links to the source literature. This design enables continuous expansion of benchmarks as new publications appear, ensuring evaluations remain timely, reproducible, and extensible.

As a case study, we apply our framework to radiation and cancer biology, generating over 16,000 MCQs from over 22,000 open-access papers and abstracts via Semantic Scholar \cite{semanticscholar}. We evaluate small and mid-sized LLMs (1.1B–14B parameters) under three settings: (i) baseline (no retrieval), (ii) retrieval-augmented generation (RAG) from paper-derived chunks, and (iii) RAG from reasoning traces produced by GPT-4.1 while answering the same questions, with final answers excluded to prevent leakage. Across models, reasoning-trace retrieval yields larger and more consistent accuracy gains than chunk retrieval. To test external validity, we repeat the comparison on an expert-written radiation and cancer biology exam. The trend persists: reasoning-trace retrieval achieves the best accuracy across all models, enabling several small models to outperform a GPT-4 baseline \cite{Beattie2024} on the 2023 ASTRO Radiation and Cancer Biology exam \cite{ASTRO2023} despite its far greater scale.

Our contributions are as follows:
\begin{enumerate}
    \item A scalable, modular pipeline for automated MCQA benchmark generation from scientific literature that is designed to utilize high-performance computing platforms.
    \item A new benchmark of 16,679 questions derived from 14115 open-access papers and 8433 abstracts in radiation and cancer biology.
    \item A systematic evaluation of 1.1B–14B parameter LLMs with RAG from both paper-derived semantic chunks and GPT-4.1 reasoning traces.
    \item Empirical results showing reasoning trace retrieval significantly and consistently improves small models towards domain , often more so than directly retrieving from literature, and in some cases leading them to exceed GPT-4 performance on expert assessments.
\end{enumerate}

This work demonstrates that scalable benchmark generation can keep pace with scientific progress while enabling smaller, more efficient models to achieve state-of-the-art performance in specialized domains. Furthermore, our framework can scale up the process of reasoning distillation from state-of-the-art Large Language Models (LLMs) for domain adaptation of Small Language Models (SLMs), rendering them capable components within agentic scientific workflows. We release our framework and benchmark artifacts to facilitate reproducible evaluation and to support future AI-for-science research. Source code and configurations for our experiments can be found in our Github repository \footnote{https://github.com/ramanathanlab/distllm}

The remainder of this paper is organized as follows. \autoref{sec:overview} introduces automated MCQA generation framework in technical details and outlines the experimental setup we designed to demonstrate its capabilities. \autoref{sec:results} discusses the empirical results of our case study on both a synthetically generated and an expert-annotated MCQA benchmark on radiation and cancer biology. \autoref{sec:related_work} briefly discusses the related work on automated scientific MCQ generation. Finally \autoref{sec:conclusion} concludes the paper and discusses future research directions.

\section{Framework and Experimental Setup}\label{sec:overview}

The framework builds on prior work in HPC-scale distributed PDF parsing \cite{siebenschuh2025adaparseadaptiveparallelpdf} and scientific RAG \cite{Gokdemir_2025}, both of which leverage Parsl \cite{babuji19parsl} for scaling on ALCF supercomputers.  

\begin{figure}[ht]
    \centering
    \includegraphics[width=8.7cm]{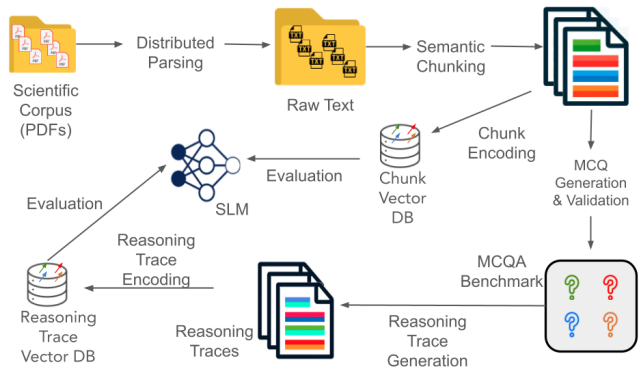}
    \caption{Workflow overview. PDFs are parsed into text and chunked semantically. Chunks are encoded (PubMedBERT) for RAG and also passed to GPT-4.1 for MCQ generation. Questions are then used for reasoning-trace generation and stored in a separate RAG database. Models are evaluated with i) no RAG, ii) chunk RAG, and iii) reasoning-trace RAG. An arbitrary LLM judge performs the grading and provides a reasoning.}
    \label{fig:tpc-main-workflow} 
\end{figure}

Our workflow begins by acquiring a scientific corpus in PDF, Markdown, or text. To demonstrate our framework with a case study, we use 14,115 full-text papers and 8,433 abstracts downloaded via the Semantic Scholar API \cite{semanticscholar} with cancer and radiation biology keywords. PDFs are parsed with AdaParse \cite{siebenschuh2025adaparseadaptiveparallelpdf}, which extracts text and metadata in JSON format. To address SLM context limits, we perform semantic chunking with PubMedBERT \cite{pubmedbert}, a 330M-parameter encoder pretrained on biomedical text, yielding 173,318 chunks.

Chunks are fed to GPT-4.1 in batches through the Argo-Proxy API \cite{ding_argoproxy}. The structured prompt first summarizes and expands the chunk, then generates an MCQ with one correct answer and distractors, while prohibiting references to the source text to ensure self-containment. A second prompt evaluates question clarity, accuracy, distractor plausibility, and educational value (score 1–10). Low-quality items are discarded; accepted items retain provenance via chunk\_id and file path.  

Each question is stored in JSON with question text, answer, original chunk, type, provenance, and metadata including relevance and quality checks.

\begin{figure}[ht]
    \centering
    \includegraphics[scale=0.6]{figures/mcq\_schema.png}
    \caption{JSON schema for generated questions. Each record contains lineage to the source chunk and PDF, along with relevance and quality checks to ensure transparent quality assurance.}
    \label{fig:mcq_schema} 
\end{figure}

This schema enables reproducibility, filtering, and provenance tracking. We generate 173,318 candidate questions (seven options each). After filtering (threshold 7/10), the benchmark contains 16,680 MCQs in radiation and cancer biology.

While this process can serve as a full end-to-end workflow, here we focus on testing whether reasoning traces from a larger model can improve smaller LLMs. To this end, semantic chunks are encoded with PubMedBERT into FP16 embeddings (747MB total) and stored in a FAISS~\cite{douze2024faiss} vector store for retrieval.  

\begin{figure}[ht]
    \centering
    \includegraphics[scale=0.5]{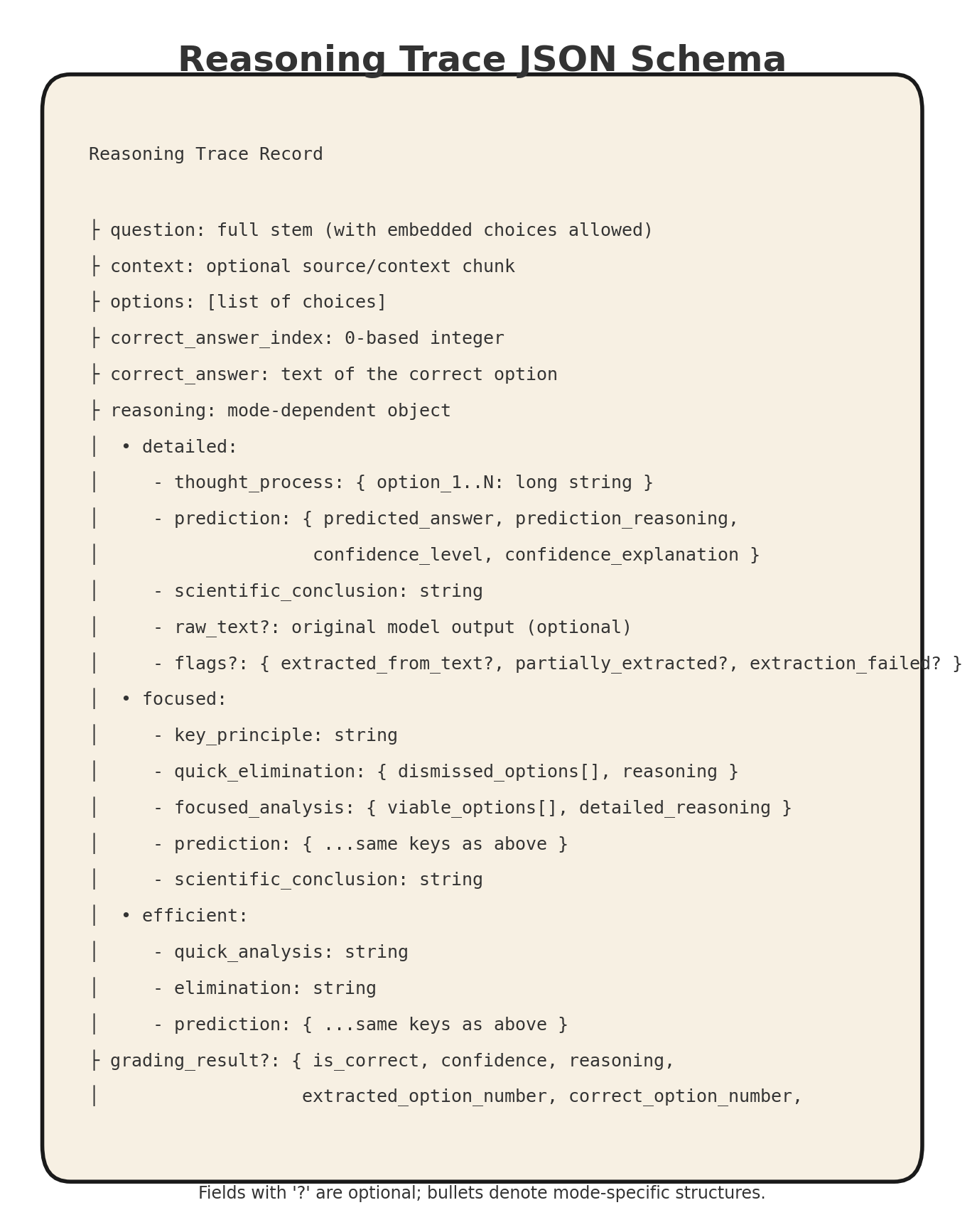}
    \caption{Reasoning-trace JSON schema. Supports three reasoning modes: detailed (option-level analysis), focused (principle + elimination), and efficient (compact high-level reasoning).}
    \label{fig:rt-json-schema} 
\end{figure}

We extract reasoning traces by prompting GPT-4.1 to answer all 16,680 MCQs, explicitly excluding the final answer to prevent leakage. Three reasoning modes are generated simultaneously (detailed, focused, efficient) and stored in separate FAISS databases for evaluation. \autoref{fig:rt-json-schema} shows the schema.  

Finally, we evaluate a suite of SLMs on both our synthetic benchmark and the 2023 ASTRO Radiation and Cancer Biology Study Guide \cite{ASTRO2023}, referred to as the Astro exam. \autoref{fig:tpc-main-workflow} illustrates the complete process.

\subsection{Selection of Small Language Models}

We evaluate a current and representative set of small and mid-sized open-source language models ranging from 1.1B to 14B parameters. The models were selected to capture the current diversity of architectures, training corpora, and licensing terms available in the open community. This range of model size was chosen for three reasons. First, our case study investigates the feasibility of domain adaptation of SLMs through reasoning distillation from retrieved reasoning traces originating from larger models. Second, as future work, we plan to investigate continual pretraining methods for domain adaptation of SLMs. Therefore, the size and weight-availability of these models render them appropriate. Finally, the relatively approachable hardware prerequisites for running these models contribute to the reproducibility of our evaluations.

We choose the following models for evaluation:
\begin{enumerate}
    \item \textbf{OLMo-7B (Allen Institute, 2024):} A 7B parameter model developed as part of the OLMo project to accelerate language model science. It supports a 2K token context window and emphasizes reproducibility.\cite{groeneveld2024olmo}
    \item \textbf{TinyLlama-1.1B-Chat (TinyLlama Team, 2024):} A compact model trained on 3T tokens with $\sim$2K context, designed as an efficient baseline for small-scale deployments. \cite{zhang2024tinyllamaopensourcesmalllanguage}
    \item \textbf{Gemma 3 4B-IT (Google, 2025):} A recent 4B parameter instruction-tuned model with a large 128K context window, representing the newest generation of mid-scale LLMs.\cite{gemmateam2025gemma3technicalreport}
    \item \textbf{SmolLM3-3B (HuggingFace, TBD):} A lightweight experimental model, evaluated to capture the behavior of smaller 3B-scale instruction-tuned systems. \cite{smollm3}
    \item \textbf{Mistral-7B-Instruct-v0.3 (Mistral AI, 2024):} A highly optimized 7B model with strong efficiency and reasoning performance, features a 4K token context size. \cite{jiang2023mistral7b}
    \item \textbf{Llama-3-8B-Instruct (Meta, 2024) and Llama-3.1-8B-Instruct (Meta, 2024):} Two successive generations of Meta’s flagship open models, included to establish strong baselines in the 8B parameter class. \cite{grattafiori2024llama3herdmodels}
    \item \textbf{Qwen-1.5-14B-Chat] (Alibaba, 2024):} A 14B parameter multilingual model with a 32K context window, representing the upper end of our evaluation range.\cite{bai2023qwentechnicalreport}
\end{enumerate}

\begin{table}[ht]
    \centering
    \caption{Overview of evaluated SLMs with parameter counts, release years, and context window sizes.}
    \resizebox{\columnwidth}{!}{
    \begin{tabular}{|l|c|c|c|}
    \hline
    \textbf{Model Name}             & \textbf{Params} & \textbf{Release Year} & \textbf{Context Window} \\
    \hline
    OLMo-7B                         & 7 B             & 2024                  & 2048 \\
    TinyLlama-1.1B-Chat             & 1.1 B           & 2024                  & 2048  \\
    Gemma 3 4B-IT                   & 4 B             & 2025                  & 128,000 \\
    SmolLM3-3B                      & 3 B             & 2025                   & 32,768 \\
    Mistral-7B-Instruct-v0.3        & 7 B             & 2024                   & 4096 \\
    Llama-3-8B-Instruct             & 8 B             & 2024          & 8192 \\
    Llama-3.1-8B-Instruct           & 8 B             & 2024         & 32,768 \\
    Qwen-1.5-14B-Chat               & 14 B            & 2024                  & 32,768 \\
    \hline
    \end{tabular}
    }
    \label{tab:slm_details}
\end{table}

\subsection{Evaluation Protocol}

Each model was tested under three conditions:
\begin{itemize}
    \item Baseline: directly prompting with the question. Off-the-shelf performance without retrieval.
    \item RAG-Chunks: retrieval-augmented generation from source paper chunks.
    \item RAG-Traces: retrieval-augmented generation from the reasoning traces of the larger model (GPT-4.1)
\end{itemize}

We report accuracy under all three conditions on both our synthetic benchmark (16,680 MCQs) and the Astro exam \cite{ASTRO2023}. The original Astro exam contains 337 questions, two of which are excluded because they require multimodal question-answering from visuals. We therefore evaluate on the remaining 335 questions in two ways: (i) cumulative accuracy across all questions, and (ii) accuracy restricted to questions that do not require mathematical reasoning or arithmetic tool use. The latter subset is identified automatically by GPT-5.

\section{Results}
\label{sec:results}

\subsection{Synthetic Benchmark Results}

\begin{figure}[ht]
    \centering
    \includegraphics[width=8.7cm]{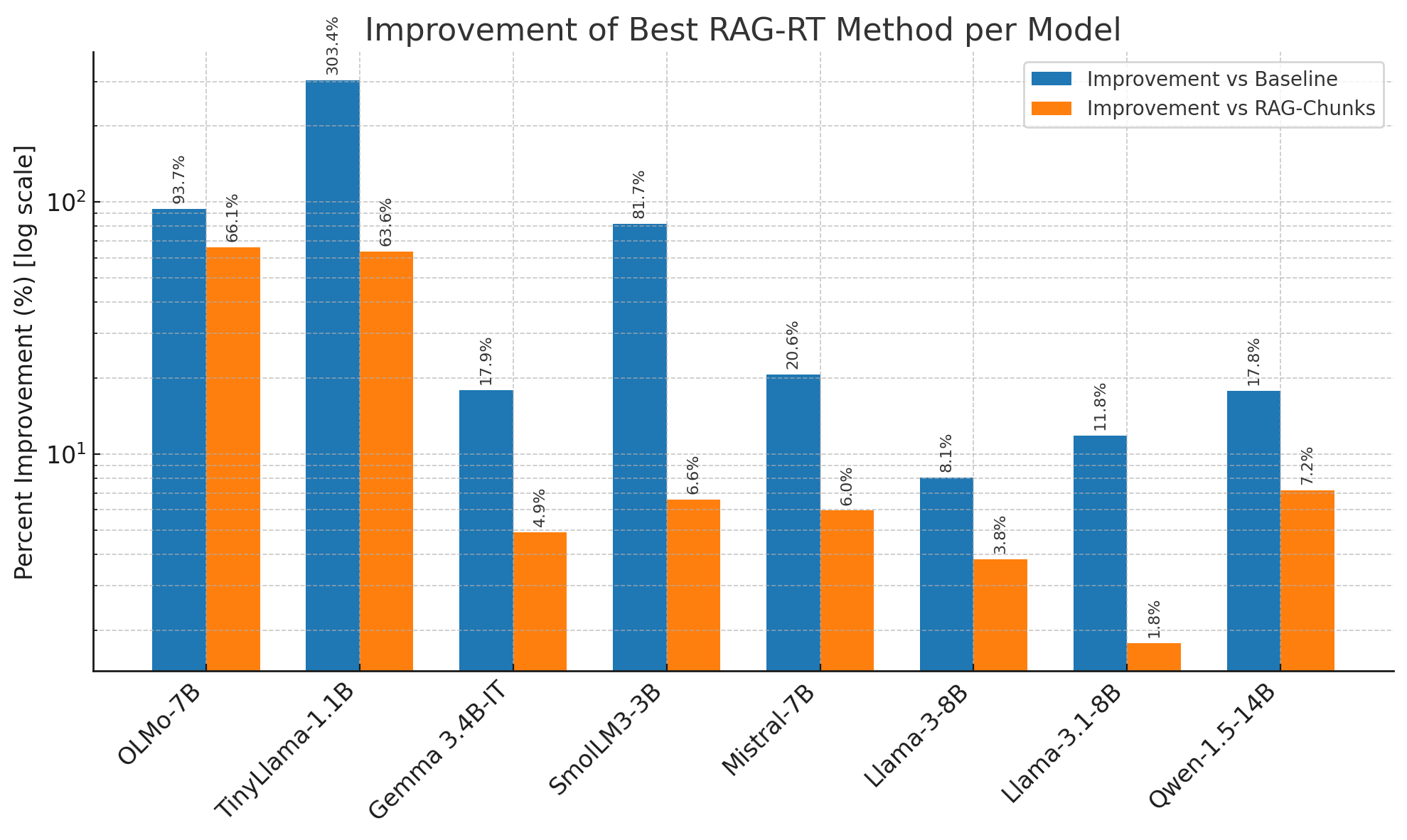}
    \caption{Percent accuracy improvement on the synthetic MCQA benchmark, comparing reasoning-trace retrieval to both baseline performance and retrieval from source documents for each evaluated model.}

    \label{fig:16k-percent-improvement} 
\end{figure}

We first evaluate models on our automatically generated benchmark of 16,680 MCQs in radiation and cancer biology. This large-scale setting allows us to systematically compare baseline performance, retrieval from source document chunks, and retrieval from reasoning traces (RAG-RT). The results are depicted in \autoref{tab:16k_eval} and \autoref{fig:16k-percent-improvement} and discussed below.

\subsubsection{Baseline vs. RAG-Chunks}

As expected, retrieval from semantic chunks provides a strong lift over baseline accuracy across nearly all models. For example, TinyLlama-1.1B-Chat improves from 17.6\% to 43.4\% accuracy, a relative gain of +147\%. Similarly, Olmo-7B moves from 38\% to 44.3\% accuracy, and Gemma-3.4B-IT increases from 74.5\% to 83.7\%. These results confirm that chunk retrieval alone provides noticeable domain adaptation for knowledge-intensive MCQA.

\begin{table}[ht]
    \centering
    \caption{Accuracy of evaluated models on the synthetic benchmark under baseline, RAG from paper chunks, and three reasoning-trace (RT) retrieval modes. Best-performing configuration per model (row) is in \textbf{bold}.}
    \resizebox{\columnwidth}{!}{
    \begin{tabular}{|l|c|c|c|c|c|}
        \hline
        \textbf{Model} & \textbf{Baseline} & \textbf{RAG-Chunks} & \textbf{RAG-RT-Detail} & \textbf{RAG-RT-Focused} & \textbf{RAG-RT-Efficient} \\
        \hline
        OLMo-7B           & 0.380 & 0.443 & 0.709 & \textbf{0.736} & 0.720 \\
        \hline
        TinyLlama-1.1B    & 0.176 & 0.434 & \textbf{0.710} & 0.699 & 0.581 \\
        \hline
        Gemma 3.4B-IT     & 0.745 & 0.837 & 0.860 & \textbf{0.878} & 0.873 \\
        \hline
        SmolLM3-3B        & 0.471 & 0.803 & 0.826 & 0.854 & \textbf{0.856} \\
        \hline
        Mistral-7B        & 0.737 & 0.839 & \textbf{0.886} & 0.889 & 0.882 \\
        \hline
        Llama-3-8B        & 0.830 & 0.864 & 0.875 & 0.892 & \textbf{0.897} \\
        \hline
        Llama-3.1-8B      & 0.819 & 0.900 & 0.915 & 0.902 & \textbf{0.916} \\
        \hline
        Qwen-1.5-14B      & 0.776 & 0.853 & 0.913 & 0.908 & \textbf{0.914} \\
        \hline
    \end{tabular}
    }
    \label{tab:16k_eval}
\end{table}

\subsubsection{RAG-RT Performance}

Reasoning-trace retrieval consistently outperforms both baseline and chunk retrieval, with the largest relative gains observed in smaller models. For example, TinyLlama-1.1B-Chat improves to 71\%–71.5\% accuracy depending on reasoning mode, nearly quadrupling its baseline performance. SmolLM3 rises from 47.1\% baseline to 85.6\% with reasoning traces (+82\% relative). Larger models such as Llama-3.1-8B and Qwen-1.5-14B-Chat also benefit, reaching peak accuracies of 91.6\% and 91.4\% respectively, though the relative gains are smaller given their stronger baseline performance.

\subsubsection{Reasoning Modes}

Across models, all three reasoning modes (detailed, focused, efficient) yield strong improvements, with only modest variation. Focused and efficient reasoning modes often provide the best balance of accuracy and retrieval efficiency, while detailed reasoning sometimes trails slightly, likely due to noise from over-elaboration. Notably, Llama-3.1-8B achieves its best performance under the efficient reasoning mode (91.6\%), suggesting that compact, distilled rationales can be just as effective as full option-by-option analyses.

\subsection{Astro Exam Results}

We next evaluate models on the 2023 ASTRO Radiation and Cancer Biology Study Guide \cite{ASTRO2023}, which contains 337 questions. Two questions were excluded for requiring multimodal capabilities. Since the SLMs that are evaluated in this chapter are not expected to have mathematical reasoning capabilities, we further divide our experiments into two settings: i) all 335 questions, and ii) the 189 questions that do not require mathematical reasoning or arithmetic tool-calling based on the classification of GPT-5 \cite{openai_gpt5_2025}.

\subsubsection{All Astro Questions}

\begin{figure}[ht]
    \centering
    \includegraphics[width=8.7cm]{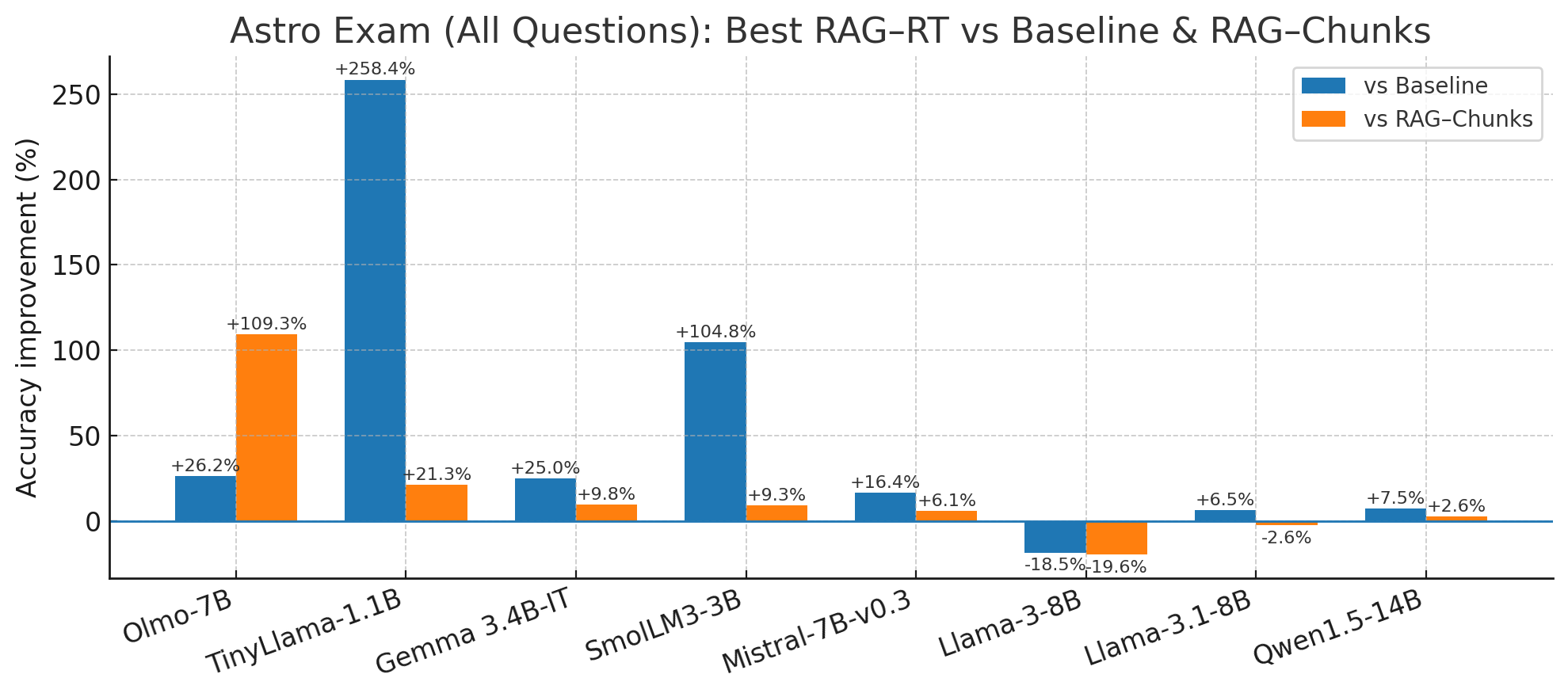}
    \caption{Percent accuracy improvement on \textit{all} questions of the 2023 ASTRO Radiation and Cancer Biology Study Guide  MCQA benchmark, comparing reasoning-trace retrieval to both baseline performance and retrieval from source documents for each evaluated model.}

    \label{fig:astro_improvement_all} 
\end{figure}

Across the full Astro exam, reasoning-trace retrieval (RAG-RT) consistently outperforms the baseline condition and usually surpasses retrieval from source document chunks. For example, Gemma-3.4B-IT improves by +25.0\% relative to baseline, while SmolLM3 and Mistral-7B-Instruct both gain ~+20\% over baseline. Even the smallest models (TinyLlama-1.1B-Chat, Olmo-7B) see gains of +15–30\% over their baseline accuracy. Notably, the relative improvements over RAG-Chunks are smaller and sometimes negative (e.g., Llama-3-8B-Instruct), suggesting that direct text retrieval already provides useful context in some cases. Nevertheless, reasoning traces remain the more stable retrieval source across models. \autoref{fig:astro_improvement_all} and \autoref{tab:astro_all} depict the empirical results.

\begin{table}[ht]
    \centering
    \caption{Astro exam (all questions): accuracy for each model under three configurations. Best per model in \textbf{bold}.}
    \resizebox{\columnwidth}{!}{
    \begin{tabular}{|l|c|c|c|}
        \hline
        Model & Baseline & RAG--Chunks & RAG--RTs (best) \\\hline
        Olmo-7B & 0.446 & 0.269 & \textbf{0.563} \\\hline
        TinyLlama-1.1B-Chat & 0.089 & 0.263 & \textbf{0.319} \\\hline
        Gemma 3.4B-IT & 0.484 & 0.551 & \textbf{0.605} \\\hline
        SmolLM3-3B & 0.377 & 0.706 & \textbf{0.772} \\\hline
        Mistral-7B-Instruct-v0.3 & 0.494 & 0.542 & \textbf{0.575} \\\hline
        Llama-3-8B-Instruct & 0.665 & \textbf{0.674} & 0.542 \\\hline
        Llama-3.1-8B-Instruct & 0.644 & \textbf{0.704} & 0.686 \\\hline
        Qwen1.5-14B-Chat & 0.560 & 0.587 & \textbf{0.602} \\\hline
    \end{tabular}
    }
    \label{tab:astro_all}
\end{table}

\subsubsection{Astro No-Math Subset}

\begin{figure}[ht]
    \centering
    \includegraphics[width=8.7cm]{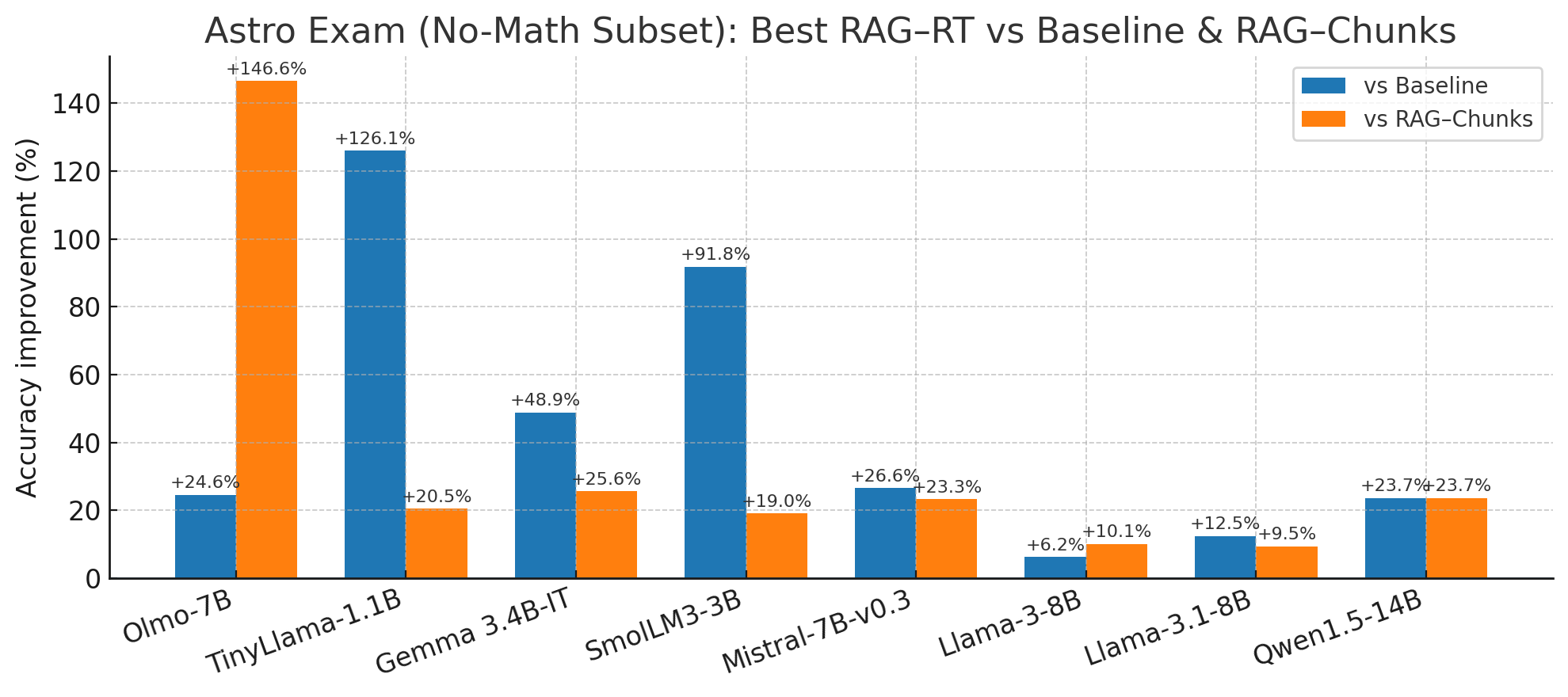}
    \caption{Percent accuracy improvement on \textit{non-mathematical} questions of the 2023 ASTRO Radiation and Cancer Biology Study Guide  MCQA benchmark, comparing reasoning-trace retrieval to both baseline performance and retrieval from source documents for each evaluated model.}

    \label{fig:astro_improvement_nomath} 
\end{figure}

When restricting to questions that do not require mathematical reasoning, the benefits of reasoning traces become more pronounced. All models show positive gains over both baseline and RAG-Chunks. SmolLM3 exhibits the largest jump, improving by nearly +92\% relative to baseline and +19\% over RAG-Chunks. Similarly, Gemma-3.4B-IT and Mistral-7B-Instruct gain +49\% and +26\% over baseline, respectively. The effect is particularly striking for smaller models (TinyLlama-1.1B-Chat, Olmo-7B), which move from near-random performance to competitive accuracy when supported by reasoning traces. This indicates that reasoning traces are especially valuable when the task does not require explicit numeric computation, aligning with the intuition that distilled scientific rationales capture high-value domain knowledge. \autoref{fig:astro_improvement_nomath} and \autoref{tab:astro_no_math} depict the results.

\begin{table}[ht]
    \centering
    \caption{Astro exam (no-math subset): accuracy for each model under three configurations. Best per model in \textbf{bold}.}
    \resizebox{\columnwidth}{!}{
    \begin{tabular}{|l|c|c|c|}
        \hline
        Model & Baseline & RAG--Chunks & RAG--RTs (best) \\\hline
        Olmo-7B & 0.471 & 0.238 & \textbf{0.587} \\\hline
        TinyLlama-1.1B-Chat & 0.138 & 0.259 & \textbf{0.312} \\\hline
        Gemma 3.4B-IT & 0.540 & 0.640 & \textbf{0.804} \\\hline
        SmolLM3-3B & 0.466 & 0.751 & \textbf{0.894} \\\hline
        Mistral-7B-Instruct-v0.3 & 0.598 & 0.614 & \textbf{0.757} \\\hline
        Llama-3-8B-Instruct & 0.757 & 0.730 & \textbf{0.804} \\\hline
        Llama-3.1-8B-Instruct & 0.762 & 0.783 & \textbf{0.857} \\\hline
        Qwen1.5-14B-Chat & 0.667 & 0.667 & \textbf{0.825} \\\hline
    \end{tabular}
    }
    \label{tab:astro_no_math}
\end{table}

\section{Related Work}
\label{sec:related_work}
Automated generation of multiple-choice question-answering (MCQA) benchmarks has been explored across domains to reduce manual curation effort. For instance, SciQ \cite{welbl2017sciq} leveraged crowdsourcing to create science MCQs, but remained constrained to preselected topics. More recently, GPQA\cite{rein2023gpqagraduatelevelgoogleproofqa} distilled high-level questions from Google’s Knowledge Graph using weak supervision, though it lacked scientific domain specificity and did not incorporate provenance. MMLU \cite{hendrycks2021mmlu} offers broad subject coverage but is largely static and vulnerable to data contamination. In biomedical domains, PubMedQA \cite{jin2019pubmedqa} generated QA pairs via abstract-based templates (Jin et al., 2019), but still relied heavily on manually crafted patterns. In astronomy, the AstroMLab-1 benchmark \cite{astroml} provides 4425 expert-vetted MCQs across astrophysics and instrumentation, but is limited to question retrieval, not generation.  Our approach extends prior work by fully automating end-to-end MCQA generation from large scientific corpora, including PDF parsing, semantic chunking, question synthesis, and quality-aware filtering, all with provenance tracking—a combination not seen in prior efforts.

\section{Conclusions and Future Work}
\label{sec:conclusion}

To keep scientific benchmarks current and domain-relevant, we develop a scalable framework for automated MCQA generation from large corpora of scientific literature. In a case study on radiation and cancer biology, the framework produces over 16,000 MCQs from 22,000 papers and abstracts and evaluates contemporary SLMs on both these synthetic questions and an expert-annotated domain exam, testing whether reasoning traces from a larger model can surpass direct retrieval from source documents.

On both the synthetic benchmark and the Astro exam, reasoning traces consistently outperform baseline and chunk retrieval. The gains are especially pronounced for smaller models, which show dramatic improvements and in some cases reach or surpass GPT-4 performance. These results demonstrate that reasoning traces distilled from a larger model provide high-value retrieval sources, substantially boosting the accuracy of smaller and mid-sized LLMs on knowledge-intensive scientific MCQA, particularly in settings where models are disadvantaged by limited training or parameter count.

Building on these results, we plan to scale our HPC-grade framework on Aurora at the Argonne Leadership Computing Facility to generate MCQ benchmarks and reasoning traces from web-scale corpora such as bioRxiv and arXiv. These benchmarks will be organized by sub-domain with metadata linking each question to its source, and we will explore pretraining LLMs on reasoning traces to systematically compare their performance against contemporary peers in scientific applications.

\section{Acknowledgement}
This material is based upon work supported by Laboratory Directed Research and Development (LDRD) funding from Argonne National Laboratory, provided by the Director, Office of Science, of the U.S. Department of Energy under Contract No. DE-AC02-06CH11357.

\bibliographystyle{ACM-Reference-Format}
\bibliography{references}

\end{document}